\documentclass[letterpaper]{article} 
\usepackage{aaai2026}  
\usepackage{times}  
\usepackage{helvet}  
\usepackage{courier}  
\usepackage[hyphens]{url}  
\usepackage{graphicx} 
\usepackage{natbib}  
\usepackage{caption} 
\usepackage{algorithm}
\usepackage{algorithmic}

\usepackage{booktabs} 
\usepackage{multirow}
\usepackage{amsmath}
\usepackage{amsfonts}

\usepackage{newfloat}
\usepackage{listings}
\DeclareCaptionStyle{ruled}{labelfont=normalfont,labelsep=colon,strut=off} 
\floatstyle{ruled}
\newfloat{listing}{tb}{lst}{}
\floatname{listing}{Listing}
\title{SpatioTemporal Difference Network for Video Depth Super-Resolution}
\author{
    Zhengxue Wang$^{1}$, Yuan Wu$^{1}$, Xiang Li$^{2}$, Zhiqiang Yan$^{3}$\thanks{Corresponding authors}, Jian Yang$^{1}$\footnotemark[1]
}
\affiliations{
    $^1$PCA Lab, Key Lab of Intelligent Perception and Systems for High-Dimensional Information of Ministry of Education, School of Computer Science and Engineering, Nanjing University of Science and Technology\\
    $^2$Nankai University  \ \
    $^3$National University of Singapore\\
    {\tt\small \{zxwang,wuyuan,csjyang\}@njust.edu.cn},  \
    {\tt\small xiang.li.implus@nankai.edu.cn, yanzq@nus.edu.sg}
}

\usepackage{bibentry}

\begin{document}

\maketitle

\begin{abstract}
Depth super-resolution has achieved impressive performance, and the incorporation of multi-frame information further enhances reconstruction quality.
Nevertheless, statistical analyses reveal that video depth super-resolution remains affected by pronounced long-tailed distributions, with the long-tailed effects primarily manifesting in spatial non-smooth regions and temporal variation zones. 
To address these challenges, we propose a novel SpatioTemporal Difference Network (STDNet) comprising two core branches: a spatial difference branch and a temporal difference branch.
In the spatial difference branch, we introduce a spatial difference mechanism to mitigate the long-tailed issues in spatial non-smooth regions. This mechanism dynamically aligns RGB features with learned spatial difference representations, enabling intra-frame RGB-D aggregation for depth calibration.
In the temporal difference branch, we further design a temporal difference strategy that preferentially propagates temporal variation information from adjacent RGB and depth frames to the current depth frame, leveraging temporal difference representations to achieve precise motion compensation in temporal long-tailed areas.
Extensive experimental results across multiple datasets demonstrate the effectiveness of our STDNet, outperforming existing approaches. 
\end{abstract}

\begin{links}
    \link{Code}{https://github.com/yanzq95/STDNet}
\end{links}

\section{Introduction}

Depth data constitutes a fundamental component in various fields, including 3D reconstruction~\cite{im2018accurate, lian2025cross, yan2025completion}, virtual reality~\cite{yan2022rignet, lian2023multi, zhou2023memory, yan2024learnable}, and augmented reality~\cite{song2020channel, yan2025event, yin2023metric3d}.
Recently, numerous depth super-resolution (DSR) methods~\cite{guo2018hierarchical, wang2023rgb, wang2023learning} have been proposed to reconstruct high-resolution (HR) depth from low-resolution (LR) inputs, achieving remarkable performance. Furthermore, \cite{sun2023consistent} introduce a video depth super-resolution (VDSR) framework that effectively aggregates multi-frame RGB-D features, demonstrating substantial improvements over single-frame approaches. 

\begin{figure}[t]
 \centering
 \includegraphics[width=0.94\columnwidth]{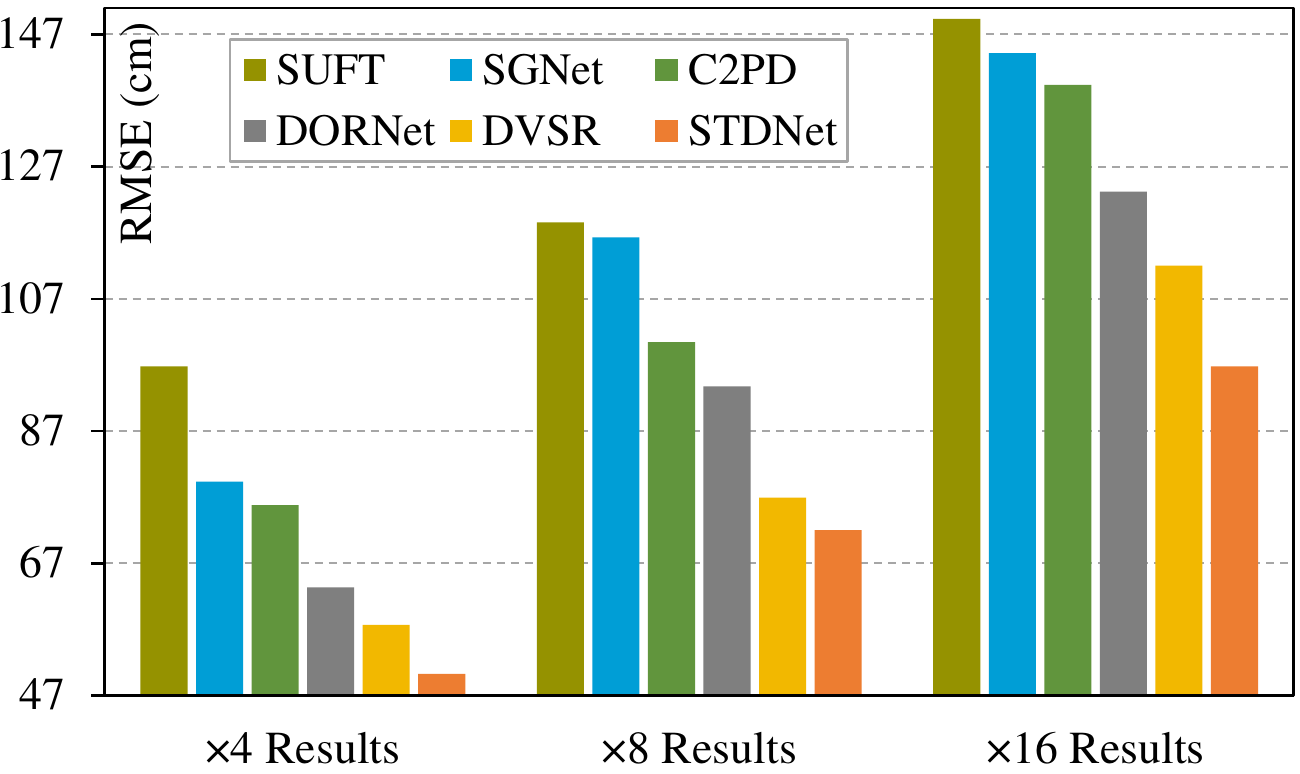}\\
 \caption{Quantitative comparisons between our STDNet and previous state-of-the-art methods on TarTanAir dataset.}
 \label{fig:radar}
\end{figure}

\begin{figure*}[t]
  \centering
    \includegraphics[width=0.96\linewidth]{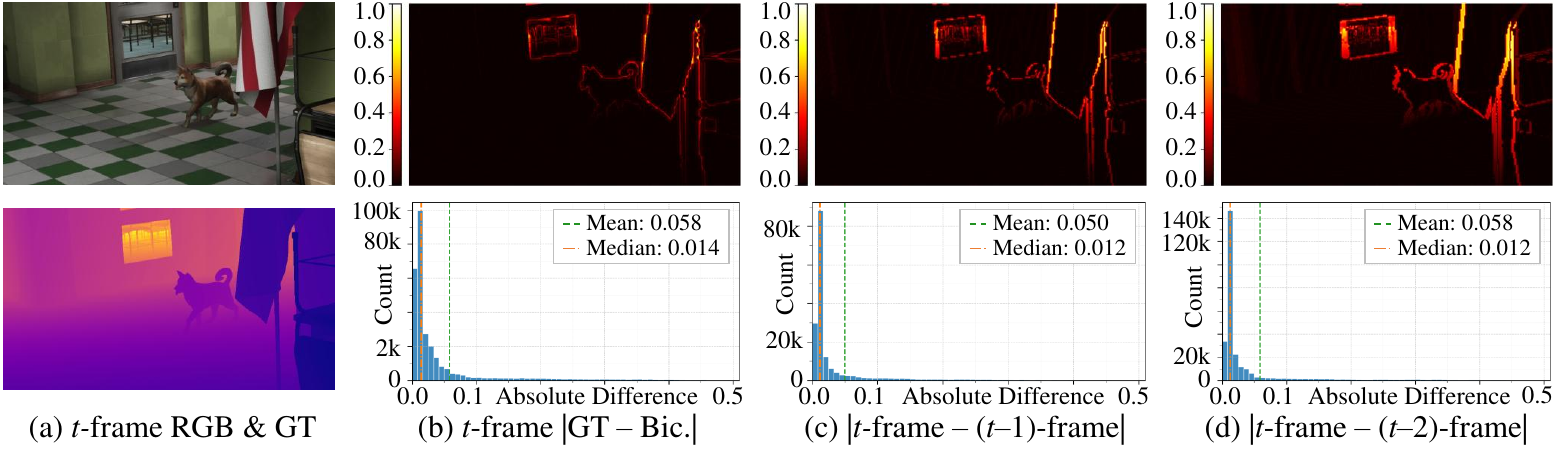}
    \caption{Visualization of (a) RGB and GT depth at frame $t$, (b) absolute difference representations (top) and corresponding histogram distribution (bottom) between GT depth and bicubic-upsampled LR depth (Bic.). (c) shows the error analysis between consecutive frames ($t$ and $t-1$ bicubic-upsampled LR depth), while (d) presents cross frame results between frames $t$ and $t-2$.}
    \label{fig:long_tailed}
\end{figure*}

However, as shown in Fig.~\ref{fig:long_tailed}, VDSR manifests long-tailed distributions across both spatial and temporal dimensions. Specifically, Fig.~\ref{fig:long_tailed}(b) quantifies the spatial difference between the ground truth (GT) depth and the upsampled LR depth. Statistical results indicate that spatial non-smooth regions exhibit distinct long-tailed issues, accounting for only a small fraction of the overall depth data. These regions pose substantially greater reconstruction challenges than the dominant smooth areas. Furthermore, Figs.~\ref{fig:long_tailed}(c) and (d) respectively present the difference results between consecutive and cross depth frames, demonstrating that temporal variation zones (e.g., dynamic objects, edge contours, and occlusion areas) in the depth videos are primarily concentrated in the long-tailed portion of the distributions.

Building upon the above statistical analysis, we propose a spatiotemporal difference network (STDNet) that focuses on handling long-tailed distributions in VDSR. The STDNet mainly consists of two dedicated branches:  spatial difference branch and temporal difference branch.
To effectively mitigate the spatial long-tailed distribution issues, our spatial difference branch implements a spatial difference mechanism. This mechanism uses learned spatial difference representations to precisely align intra-frame RGB features with non-smooth depth regions. These aligned RGB features are selectively aggregated to enhance the depth prediction.
In the temporal difference branch, we prioritize regions with significant temporal variations, which typically exhibit long-tailed characteristics. To this end, we first estimate the temporal difference representations between consecutive frames and cross frames in the depth videos. Then, we develop a temporal difference strategy that propagates information from multiple adjacent RGB and depth frames to the current depth frame. In this strategy, spatial and temporal difference representations are jointly employed to facilitate the alignment of multi-frame and multi-modal RGB-D data, enhancing temporal consistency in the predicted depth videos. 
Furthermore, we introduce a difference regularization comprising both spatial and temporal difference terms to optimize the learning of spatiotemporal difference representations.

Owing to these innovative designs, our method successfully restores accurate HR depth videos. As shown in Fig.~\ref{fig:radar}, 
STDNet outperforms five state-of-the-art approaches by 32.6\% ($\times4$), 28.8\% ($\times8$), and 27.6\% ($\times16$) in average.

In summary, our contributions are as follows:

\begin{itemize}
    \item Based on statistical analysis, we introduce a novel VDSR perspective that exploits the spatiotemporal long-tailed characteristics to enhance depth videos.
    \item We propose a novel framework termed STDNet, which comprises dual spatiotemporal difference branches. The spatial difference branch focuses on mitigating the long-tailed effects in spatial non-smooth regions, while the temporal difference branch prioritizes multi-frame RGB-D aggregation in temporal variation areas.
    \item Extensive experiments demonstrate that our STDNet effectively recovers high-quality depth videos, achieving state-of-the-art performance.
\end{itemize}

\begin{figure*}[t]
  \centering
    \includegraphics[width=0.9\linewidth]{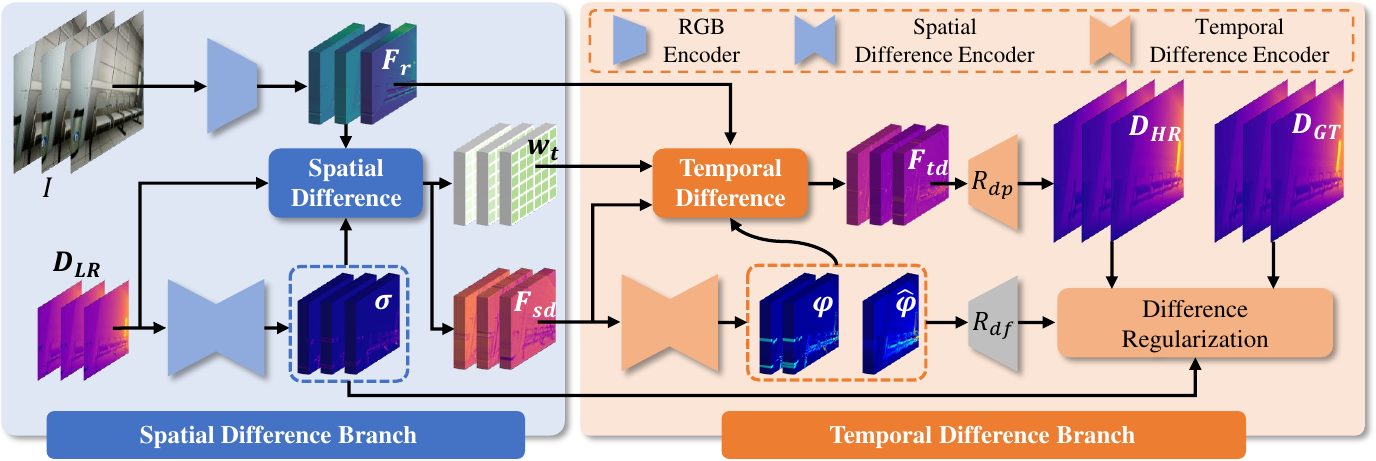}
    \caption{Overview of STDNet. Given $\boldsymbol D_{LR}$, we first predict its spatial difference representation $\boldsymbol \sigma $. Then, $\boldsymbol D_{LR}$, $\boldsymbol I$, and $\boldsymbol \sigma $ are jointly fed into the spatial difference to enhance non-smooth regions, producing $\boldsymbol F_{sd}$. Next, we estimate the temporal difference representations for consecutive frames and cross frames, generating $\boldsymbol \varphi $ and $\widehat{\boldsymbol \varphi} $. These difference representations are used to propagate adjacent RGB and depth frames to the current depth frame, generating HR depth video $\boldsymbol D_{HR}$. Finally, a degradation regularization takes $\boldsymbol D_{HR}$, $\boldsymbol D_{GT}$, $\boldsymbol \sigma $, $\boldsymbol \varphi $, and $\widehat{\boldsymbol \varphi} $ as inputs to optimize the learning of spatiotemporal difference representations.
    }
    \label{fig:Pipline}
\end{figure*} 

\section{Related Work}

\subsection{Depth Super-Resolution}
Recently, single-frame DSR methods~\cite{ye2020pmbanet, de2022learning, chen2024intrinsic} have made remarkable progress.  Existing approaches can be broadly categorized into filtering-based methods~\cite{metzger2023guided, zhong2023deep, wang2024scene}, multi-modal fusion-based methods~\cite{zhong2021high, wang2022depth, zhao2023spherical}, multi-task collaborative methods~\cite{sun2021learning, yan2022learning}, and  structure-oriented methods~\cite{yan2025ducos, bi2025dual, zheng2025decoupling, bi2025structure}. For example, \cite{kim2021deformable} integrate deformable networks with joint image filtering to adaptively transfer RGB information to depth features. To effectively aggregate RGB-D, \cite{zhao2022discrete} introduce a discrete cosine network to disentangle the shared and private information present in RGB and depth features. Additionally, \cite{sun2021learning} and \cite{tang2021bridgenet} employ an auxiliary depth estimation network to effectively fuse RGB and depth features. More recently, several methods have focused on reconstructing high-frequency information. For instance, \cite{yuan2023recurrent} develop a recurrent structure attention to perform frequency decomposition and edge restoration. Unlike these single-frame approaches, our method focuses more on addressing the spatiotemporal long-tailed distribution issues inherent in VDSR,  enabling robust reconstruction of temporally consistent HR depth videos.

\subsection{Video RGB Super-Resolution}

Single-modal video RGB super-resolution (VSR) aims to restore HR RGB videos from the corresponding LR inputs. Existing methods~\cite{isobe2022look, gao2022lightweight, shi2022rethinking} can be categorized into sliding window-based and recurrent-based approaches. Specifically, sliding window-based methods~\cite{wang2019edvr, li2020mucan, cao2021video} employ a temporal sliding window to align neighboring frames with the current frame within the window, achieving impressive performance. However, such methods are typically limited by their fixed window size and can only integrate information from a restricted number of nearby frames. To overcome this constraint, recent work has introduced bidirectional recurrent architectures~\cite{chan2021basicvsr, hu2025exploiting, zhou2024video} that can aggregate information across the entire video sequence for feature enhancement. In contrast to VSR, VDSR presents unique challenges in establishing multi-frame and multi-modal correspondences between RGB and depth videos. These fundamental discrepancy hinder the effectiveness of existing advanced VSR methods in reconstructing depth videos.

\subsection{Video Depth Enhancement}

Although single-frame depth enhancement methods have demonstrated success in recovering high-quality depth from degraded inputs, they often exhibit suboptimal performance and temporal inconsistency when applied to depth video sequences. To address these limitations, recent advances in video depth enhancement have investigated temporal fusion strategies that integrate information from both current and adjacent frames. For example, \cite{sun2023consistent} propose the first dToF-based VDSR method, which effectively harnesses rich information from multi-frame videos sequences to mitigate spatial ambiguity in LR depth videos. In addition, \cite{dong2024exploiting} develop a multi-frame depth denoising network that models the intra-scene geometric correlations and inter-frame noise distribution correlations, effectively suppressing multi-path interference and shot noise. More recently, \cite{zhu2025svdc} introduce a video depth completion framework that integrates multi-frame features through an adaptive frequency selection fusion module.  Different from them, we focus on leveraging the long-tailed distribution characteristics of depth videos to enhance both non-smooth regions and temporal variation zones.

\section{Method}

\subsection{Overall Architecture}
As illustrated in Fig.~\ref{fig:Pipline}, our STDNet mainly comprises two branches: a spatial difference branch and a temporal difference branch. In the spatial difference branch, we first predict spatial difference representation $\boldsymbol \sigma \in \mathbb{R} ^{T\times h\times w\times c} $ from the LR depth video $\boldsymbol D_{LR} \in \mathbb{R} ^{T\times h\times w\times 1} $. $c$, $h$, $w$, and $T$ are the channels, height, width, and the number of frames respectively. Then, $\boldsymbol D_{LR}$, $\boldsymbol \sigma$, and RGB video $\boldsymbol I \in \mathbb{R} ^{T\times sh\times sw\times 3} $ are fed into the spatial difference, which adaptively performs intra-frame RGB-D aggregation to enhance spatial non-smooth of LR depth video, producing enhanced depth feature $\boldsymbol F_{sd}$ and spatial difference weights $\boldsymbol w_{t} $. $s$ is upsampling factor. In the temporal difference branch, we utilize $\boldsymbol F_{sd}$ to estimate temporal difference representations for both consecutive and cross frames, yielding  $\boldsymbol \varphi  \in \mathbb{R} ^{(T-1)\times h\times w\times c}$ and $\widehat{\boldsymbol \varphi}  \in \mathbb{R} ^{(T-2)\times h\times w\times c}$, respectively. Subsequently, the temporal difference takes $\boldsymbol F_{sd}$, $\boldsymbol w_{t}$, $\boldsymbol \varphi$, $\widehat{\boldsymbol \varphi}$, and RGB feature $\boldsymbol F_{r}$ as inputs to selectively transform adjacent RGB-D frames into current depth frame, optimizing the temporal variation zones in the depth video. Finally, the reconstructed HR depth video $\boldsymbol D_{HR} \in \mathbb{R} ^{T\times sh\times sw\times 1} $, GT depth video $\boldsymbol D_{GT} \in \mathbb{R} ^{T\times sh\times sw\times 1} $, and spatiotemporal difference representations ($\boldsymbol \sigma$, $\boldsymbol \varphi $, and $\widehat{\boldsymbol \varphi} $) are input into the difference regularization to facilitate difference learning.  

\begin{figure*}[t]
 \centering
 \includegraphics[width=0.9\linewidth]{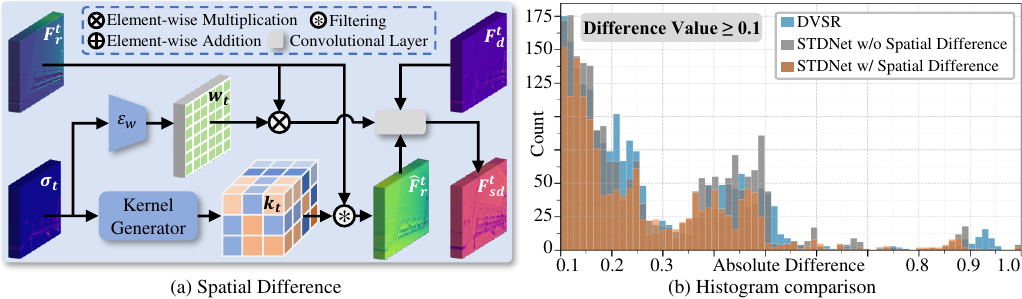}\\
 \caption{Details of (a) spatial difference, and (b) histogram comparison between our STDNet and DVSR~\cite{sun2023consistent}.}
 \label{fig:sd_hist}
\end{figure*}

\subsection{Spatial Difference Branch}
As shown in the blue part of Fig.~\ref{fig:Pipline}, our spatial difference branch is designed to accurately enhance spatial non-smooth regions. Specifically, we first encode LR depth video $\boldsymbol D_{LR}$ to depth feature $\boldsymbol F_{d}$, and then predict its spatial difference representation $\boldsymbol \sigma$:
\begin{equation}
   \boldsymbol \sigma = |\boldsymbol F_{d} - f_{bu} (f_{bd} (\boldsymbol F_{d}))|,
\end{equation}
where $f_{bd}$ and $f_{bu}$ are bilinear downsampling and upsampling operations ($\times2$), respectively. Subsequently, the RGB feature $\boldsymbol F_{r}$, depth feature $\boldsymbol F_{d}$, and $\boldsymbol \sigma$ are fed into the proposed spatial difference for intra-frame RGB-D aggregation.

\noindent \textbf{Spatial Difference.} Our spatial difference is depicted in the Fig.~\ref{fig:sd_hist}(a). Given $t$-th frame difference representation $\boldsymbol \sigma_{t}$, we first generate filtering kernel $\boldsymbol k_{t}$ through a kernel generator $\mathcal{G} $, which will be applied to filter the RGB features, ensuring their alignment with spatial non-smooth regions:
\begin{equation}
   \boldsymbol k_{t} = \mathcal{G} (\boldsymbol \sigma_{t}),
\end{equation}
where generator $\mathcal{G}$ is composed of multiple convolutional layers and activation function layers.


Then, we employ encoder $\mathcal{E} _{w}$ transforms the $\boldsymbol \sigma_{t}$ into adaptive weight $\boldsymbol w_{t}$:
\begin{equation}
   \boldsymbol w_{t}=\mathcal{E} _{w} (\boldsymbol \sigma_{t}),
\end{equation}
where $\mathcal{E} _{w}$ consists of a $3\times3$ convolutional layer, maximum function, mean  function, and sigmoid function.

Finally, the predicted filtering kernel $\boldsymbol k_{t}$ and weight $\boldsymbol w_{t}$ are leveraged to selectively propagate the aligned $t$-th frame RGB features to the depth feature $\boldsymbol F_{d}^{t}$, yielding the calibrated $t$-th frame depth feature $\boldsymbol F_{sd}^{t}$:
\begin{equation}
   \boldsymbol F_{sd}^{t} = f_{c}(\boldsymbol F_{d}^{t},\boldsymbol w_{t}\otimes \boldsymbol F_{r}^{t}, \widehat{\boldsymbol F} _{r}^{t}),
\end{equation}
where $\widehat{\boldsymbol F} _{r}^{t} =\mathcal{F} (\boldsymbol F_{r}^{t},\boldsymbol k_{t})$. $f_{c}$ is a convolutional layer, while $\otimes$ and $\mathcal{F}$ are the element-wise multiplication and filtering operation, respectively. Fig.~\ref{fig:sd_hist}(b) compares the histogram distributions in the long-tailed regions (difference value $\ge $ 0.1). Compared to state-of-the-art DVSR,  our spatial difference effectively mitigates spatial long-tailed issues, thereby enhancing non-smooth regions in LR depth videos.

\subsection{Temporal Difference Branch}

Statistical results reveal that depth videos exhibit significant long-tailed distributions along the temporal dimension. Unlike existing multi-frame fusion methods, our temporal difference strategy prioritizes motion compensation in temporal variation regions. As illustrated in the orange part of Fig.~\ref{fig:Pipline}, we first employ a temporal difference encoder to predict two difference representations: $\boldsymbol \varphi $ for consecutive frames and $\widehat{\boldsymbol \varphi} $ for cross frames:
\begin{equation}\label{eq:dc}
\begin{split}
    &\boldsymbol \varphi_{t} =|\boldsymbol F_{sd}^{t} -\boldsymbol F_{sd}^{t+1}|, \forall t\in\{ 1,2,\cdots ,T-1  \},  \\
    &\widehat{\boldsymbol \varphi}_{t}  =|\boldsymbol F_{sd}^{t} -\boldsymbol F_{sd}^{t+2}|, \forall t\in\{ 1,2,\cdots ,T-2  \} .
\end{split}
\end{equation}

Given $\boldsymbol F_{r}$, $\boldsymbol w_{t}$, $\boldsymbol F_{sd}$, $\boldsymbol \varphi $ and $\widehat{\boldsymbol \varphi} $, we then follow a common bidirectional iterative scheme~\cite{chan2022basicvsr++, sun2023consistent} to execute the proposed temporal difference strategy, fully aggregating information from adjacent RGB and depth frames. Please see our appendix for more iterative details.


Next, all forward and backward output features are aggregated to generate the temporally enhanced depth feature sequence $\boldsymbol F_{td}$. Finally, the HR depth videos are reconstructed through the depth reconstruction module $\mathcal{R} _{dp} $, which consists of convolutional layers and pixel shuffle layers:
\begin{equation}
   \boldsymbol D_{HR}=\mathcal{R}_{dp} (\boldsymbol F_{td}).
\end{equation}

\begin{figure*}[t]
 \centering
 \includegraphics[width=0.9\linewidth]{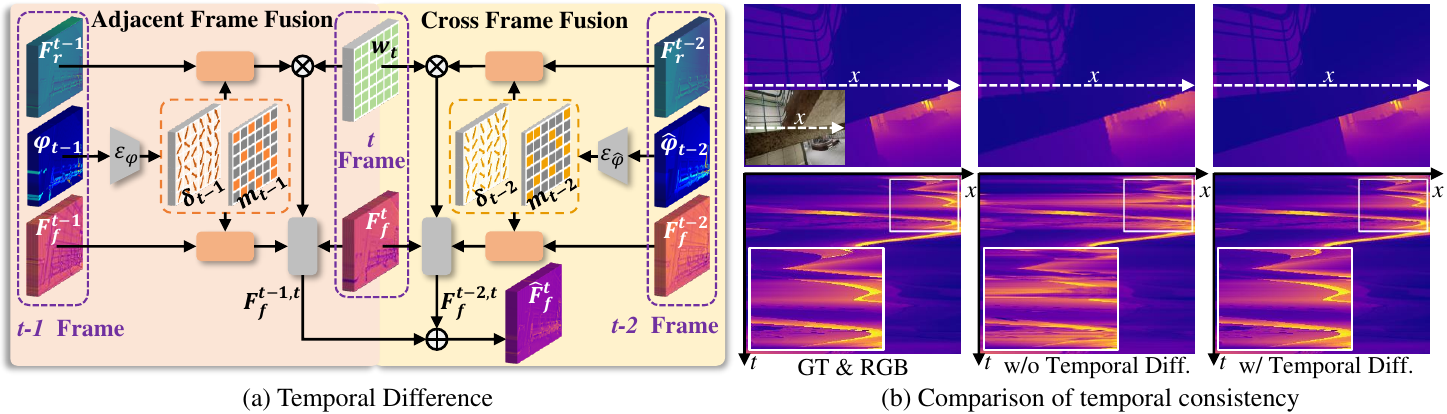}\\
 \caption{Details of (a) temporal difference, and (b) temporal consistency visualization for $x$–$t$ slices (along dashed line). Diff.: Difference. Orange rectangular boxes are the deformable convolutional layers~\cite{zhu2019deformable}.}
 \label{fig:td_hist}
\end{figure*}

\begin{table*}[t]
\centering
\footnotesize
\renewcommand\arraystretch{1.05}
\setlength{\tabcolsep}{2.5mm}{
\begin{tabular}{l|ccccccccc|c}
\toprule 
\multirow{2}{*}{Methods}  &\multicolumn{3}{c}{$\times4$}   &\multicolumn{3}{c}{$\times8$}  &\multicolumn{3}{c|}{$\times16$} &\multirow{2}{*}{Venue} \\ 
\cmidrule(lr){2-4}\cmidrule(lr){5-7}\cmidrule(lr){8-10}
 &RMSE$\downarrow $ &MAE$\downarrow $ &TEPE$\downarrow $     &RMSE$\downarrow $ &MAE$\downarrow $ &TEPE$\downarrow $  
 &RMSE$\downarrow $ &MAE$\downarrow $ &TEPE$\downarrow $  &   \\ \midrule

DJFR               &75.56  &10.59  &10.19                  &105.45 &18.43  &14.15            &141.14 &31.22  &20.27  &PAMI 2019   \\
CUNet              &89.38  &14.11  &11.64                  &122.56 &22.82  &15.93            &155.00 &38.56  &21.30  &PAMI 2020   \\
DKN                &82.69  &11.73  &10.83                  &110.10 &18.78  &14.49            &153.56 &33.21  &21.93  &IJCV 2021   \\
FDKN               &79.39  &11.14  &10.66                  &109.10 &18.48  &14.79            &147.61 &29.31  &19.77  &IJCV 2021   \\
FDSR               &80.18  &13.34  &11.79                  &104.77 &19.12  &14.52            &132.52 &29.09  &19.28  &CVPR 2021   \\
SUFT               &96.80  &15.87  &11.93                  &118.57 &22.09  &15.45            &149.40 &34.72  &19.46  &MM 2022   \\            
SGNet              &79.40  &11.36  &9.40                   &116.33 &23.15  &12.83            &144.17 &34.34  &20.14  &AAAI 2024   \\  
C2PD               &75.83  &13.12  &10.36                  &100.48 &18.70  &13.03            &139.36 &40.86  &19.32  &AAAI 2025   \\ 
DORNet             &63.38  &8.60   &8.00                   &93.75  &13.96  &11.90            &123.24 &23.59  &16.40  &CVPR 2025   \\
DVSR               &\underline{57.72}  &\underline{4.40}   &\underline{5.33}                   &\underline{76.96}  &\underline{7.74}   &\underline{8.04}     &\underline{112.04} &\underline{14.39}  &\underline{11.06}   &CVPR 2023  \\ 
\textbf{STDNet}    &\textbf{50.28}  &\textbf{3.73}   &\textbf{4.58}                   &\textbf{72.03}  &\textbf{6.75}   &\textbf{6.54}   &\textbf{96.80}  &\textbf{12.01}  &\textbf{8.90}    &-  \\
\bottomrule
\end{tabular}}
\caption{Quantitative comparisons with existing state-of-the-art methods on the TarTanAir dataset.}\label{tab:tanair}
\end{table*}

\noindent \textbf{Temporal Difference.} Our temporal difference is delineated in Fig.~\ref{fig:td_hist}(a), consists of adjacent frame fusion and cross frame fusion. Taking forward propagation as an example, we use RGB and depth features from adjacent frames ($t-1$ and $t-2$) to enhance temporal variation regions in the $t$-th frame depth, guided by temporal difference representations $\boldsymbol \varphi_{t-1} $ and $\widehat{\boldsymbol \varphi}_{t-2} $. Conversely, the backward propagation employs subsequent RGB-D frames ($t+1$ and $t+2$) for complementary refinement of $t$-th frame depth.

In the adjacent frame fusion stage, we utilize a temporal difference encoder $\mathcal{E} _{\varphi } $ (composed of convolutional layers and sigmoid function) to project $\boldsymbol {\varphi}_{t-1}$ into offset $\delta _{t-1} $ and modulation scalar $m _{t-1} $. Deformable convolution $\mathcal{D}$~\cite{zhu2019deformable} is then introduced to dynamically sample temporal variation information matched with the temporal difference representations. Additionally, adaptive weight $\boldsymbol w_{t}$ derived from the spatial difference branch is employed to  mitigate cross-modal discrepancies between adjacent RGB frames and the current depth frames. The enhanced depth feature $\boldsymbol F_{f}^{t-1,t}$ for frame $t$ can be expressed as:
\begin{equation}
   \boldsymbol F_{f}^{t-1,t}=f_{c} (\boldsymbol F_{f}^{t}, \boldsymbol F_{f,dc}^{t-1}, \boldsymbol w_{t}\otimes\boldsymbol F_{r,dc}^{t-1}) ,
\end{equation}
where intermediate feature $\boldsymbol F_{f,dc}^{t-1}=\mathcal{D}(\boldsymbol F_{f}^{t-1},\delta _{t-1},m_{t-1})$, $\boldsymbol F_{r,dc}^{t-1}=\mathcal{D}(\boldsymbol F_{r}^{t-1},\delta _{t-1},m_{t-1})$. $\boldsymbol F_{f}^{t}$ and $\boldsymbol F_{f}^{t-1}$ are the depth features of frame $t$ and frame $t-1$ during forward iteration.



The cross frame fusion stage follows an analogous procedure to produce feature $\boldsymbol F_{f}^{t-2,t}$. Finally, the temporally-refined feature $\widehat{\boldsymbol F}_{f}^{t}$ for frame $t$ are generated through the integration of adjacent frame fusion and cross frame fusion:
\begin{equation}
   \widehat{\boldsymbol F}_{f}^{t}=\boldsymbol F_{f}^{t-1,t}+\boldsymbol F_{f}^{t-2,t}.
\end{equation}

Fig.~\ref{fig:td_hist}(b) visualizes central slices (white dashed line) across all frames of the predicted depth videos, demonstrating that our temporal difference strategy contributes to more stable temporal predictions, particularly in regions with temporal variations that exhibit long-tailed distributions. These results confirm that STDNet effectively enhances the temporal consistency of depth videos.

\begin{figure*}[t]
  \centering
    \includegraphics[width=0.9\linewidth]{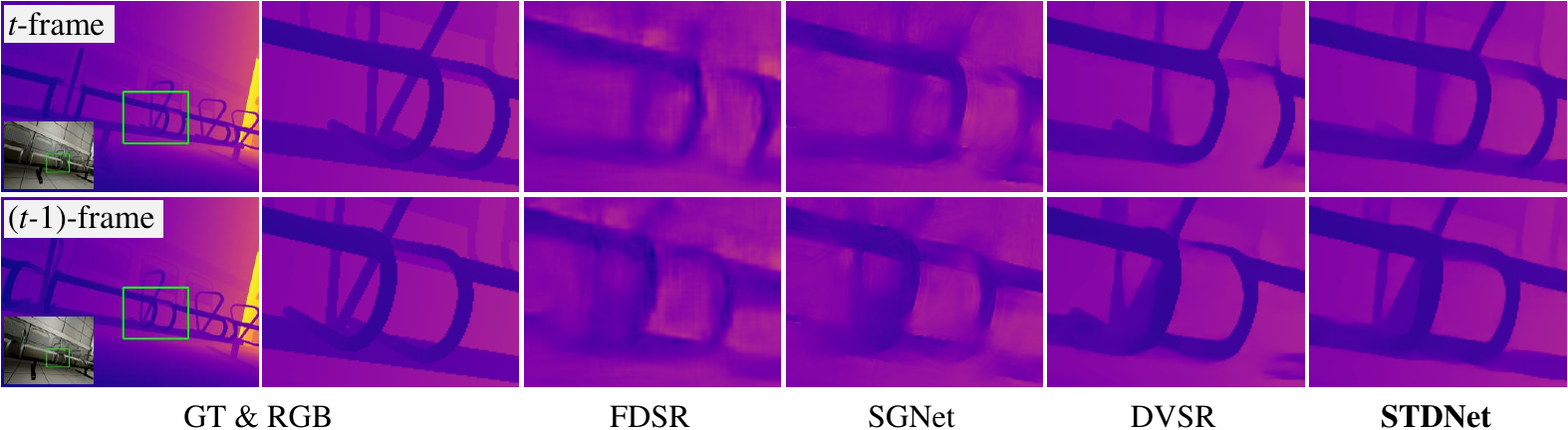}
    \caption{Visual results of consecutive frames on the TarTanAir at $\times16$ upscaling.}
    \label{fig:TarTanAir_v}
\end{figure*} 

\begin{figure*}[t]
  \centering
    \includegraphics[width=0.9\linewidth]{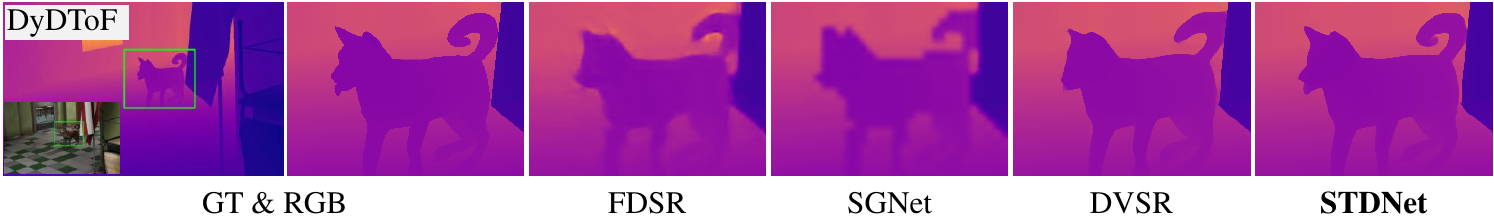}
    \caption{Visual results on the DyDToF at $\times8$ upscaling.}
    \label{fig:DyDToF_v}
\end{figure*} 

\subsection{Loss Function}
Given the GT depth video $\boldsymbol D_{GT}$ and predicted HR depth video $\boldsymbol D_{HR}$, we follow prior work~\cite{sun2023consistent} by applying Charbonnier regularization~\cite{charbonnier1994two} as reconstruction loss  $\mathcal{L} _{rec}$ to constrain our STDNet: 
\begin{equation}
   \mathcal{L} _{rec} =  {\textstyle \sum_{q\in \mathbb{Q}}}\sqrt{(\boldsymbol D_{GT}^{q} -  \boldsymbol D_{HR}^{q})^{2} +\epsilon} ,
\end{equation}
where $\mathbb{Q}$ is the set of valid pixels of $\boldsymbol D_{GT}$. $\epsilon=1\times 10^{-12} $.

\noindent \textbf{Degradation Regularization.} To optimize the learned spatiotemporal difference representations, we introduce difference regularization $\mathcal{L} _{diff}$ comprising two terms: spatial difference term $\mathcal{L} _{sd}$ and temporal difference term $\mathcal{L} _{td}$:
\begin{equation}
   \mathcal{L} _{diff} =\alpha _{1} \mathcal{L} _{sd} + \alpha _{2} \mathcal{L} _{td}.
\end{equation}
where $\alpha _{1} $ and $\alpha _{2}$ are tunable hyper-parameters.

For the spatial difference term, we introduce an uncertainty constraint~\cite{ning2021uncertainty} to facilitate the learning of difference representations in non-smooth regions:
\begin{equation}
   \mathcal{L} _{sd} = {\textstyle \sum_{q\in \mathbb{Q}}}(\boldsymbol \sigma^{q} -min(\boldsymbol \sigma^{q} ))||\boldsymbol D_{GT}^{q} -  \boldsymbol D_{HR}^{q}||_{1},
\end{equation}
where $||\cdot ||_{1} $ represents the $L_{1}$ norm.

The temporal difference term consists of two components: adjacent frame and cross frame losses:
\begin{equation}
\begin{split}
    &\mathcal{L} _{td} =\underbrace{ {\textstyle \sum_{q\in \mathbb{Q}}} ||\mathcal{R}_{df}(\boldsymbol \varphi^{q} )-\Phi  (\boldsymbol D_{GT}^{q})||_{1} }_{\text{adjacent frame}} \\
    &+ \underbrace{ {\textstyle \sum_{q\in \mathbb{Q}}}||\mathcal{R}_{df}(\widehat{\boldsymbol \varphi} ^{q} )-\Phi  (\boldsymbol D_{GT}^{q})||_{1} }_{\text{cross frame}} ,
    \label{eq_loss_lagrangian}
\end{split}
\end{equation}
where $\mathcal{R}_{df}$ represents the difference reconstruction, composed of bicubic upsampling and convolutional layers. $\Phi $ is temporal difference computation, as defined in Eq.~\eqref{eq:dc}.

The total loss $\mathcal{L} _{total}$ integrates both reconstruction loss $\mathcal{L} _{rec}$ and difference regularization $\mathcal{L} _{diff}$, formulated as:
\begin{equation}
   \mathcal{L} _{total} =\mathcal{L} _{rec} + \beta \mathcal{L} _{diff}.
\end{equation}
where $\beta$ is a tunable hyper-parameter.

\section{Experiments}
\subsection{Experimental Setups}

\noindent \textbf{Datasets.} Following previous methods~\cite{sun2023consistent, zhu2025svdc}, we evaluate STDNet on TarTanAir~\cite{wang2020TarTanAir}, DyDToF~\cite{sun2023consistent}, and DynamicReplica~\cite{karaev2023dynamicstereo} datasets. Since dataset pre-processing details from prior approaches are unavailable, we redefine the training and test sets. Specifically, we utilize the hard scenes from TarTanAir for training, consisting of $6,164$ RGB-D frames in the train set and $1,228$ frames in the test set. Then, the pre-trained model on TarTanAir is evaluated on DyDToF ($576$ frames) and DynamicReplica ($500$ frames) without any fine-tuning. Besides, DynamicReplica is center-cropped to match the size of TarTanAir ($640\times480$). All LR depth are generated from GT and RGB using the same synthesis pipeline as DVSR~\cite{sun2023consistent}. We retrain the existing methods from scratch using their released code.

\noindent \textbf{Implementation Details.} Consistent with prior works~\cite{sun2023consistent, zhu2025svdc}, we adopt root mean square error (RMSE), mean absolute error (MAE), and  temporal end-point error (TEPE) as evaluation metrics (all measured in centimeters). During training, we randomly crop RGB and HR depth frames to $256\times256$. Besides, we employ the Adam optimizer~\cite{kingma2014adam} with an initial learning rate of $1\times 10^{-4} $ to train our STDNet. Our method is implemented using two NVIDIA RTX 4090 GPUs. The hyper-parameters are set as $\alpha _{1}$=$\alpha _{2}$=0.5 and  $\beta$=0.01.

\subsection{Comparison with the State-of-the-Art}
We compare our STDNet with existing state-of-the-art approaches, including DJFR \cite{li2019joint}, CUNet \cite{deng2020deep}, DKN \cite{kim2021deformable}, FDSR \cite{he2021towards}, SUFT \cite{shi2022symmetric}, SGNet \cite{wang2024sgnet}, C2PD \cite{kang2025c2pd}, DORNet \cite{wang2025dornet}, and DVSR \cite{sun2023consistent}.

\begin{table}[t]
\centering
\small
\renewcommand\arraystretch{1.05}
\setlength{\tabcolsep}{3pt}{
\begin{tabular}{l|cccccc}
\toprule 
\multirow{2}{*}{Methods}   &\multicolumn{2}{c}{$\times4$}   &\multicolumn{2}{c}{$\times8$}  &\multicolumn{2}{c}{$\times16$} \\ 
\cmidrule(lr){2-3}\cmidrule(lr){4-5}\cmidrule(lr){6-7}
&RMSE$\downarrow $ &MAE$\downarrow $     &RMSE$\downarrow $ &MAE$\downarrow $  
 &RMSE$\downarrow $ &MAE$\downarrow $    \\ \midrule

DJFR                                         &22.39  &4.72                         &32.26  &6.90               &44.76  &11.73        \\
CUNet                                        &26.46  &6.19                         &37.74  &14.45              &54.73  &32.72        \\
DKN                                          &23.32  &4.81                         &32.20  &6.91               &48.55  &11.53        \\
FDKN                                         &22.61  &4.65                         &32.17  &7.10               &43.82  &11.57        \\
FDSR                                         &22.77  &5.20                         &31.71  &7.75               &47.86  &14.96        \\
SUFT                                         &59.06  &17.37                        &63.00  &20.35              &80.35  &36.97        \\            
SGNet                                        &40.62  &11.24                        &51.24  &13.93              &64.51  &21.74       \\  
C2PD                                         &26.05  &6.35                         &31.19  &11.68              &49.12  &29.03       \\ 
DORNet                                       &30.50  &7.07                         &54.13  &16.03              &60.27  &22.86       \\ 
DVSR                                         &\underline{19.53}  &\underline{3.16}                   &\underline{27.63}  &\underline{4.37}         &\underline{43.55}  &\underline{9.80}       \\ 
\textbf{STDNet}                              &\textbf{18.23}  &\textbf{3.04}                         &\textbf{26.87}  &\textbf{4.09}               &\textbf{39.24}  &\textbf{8.72}        \\
\bottomrule
\end{tabular}}
\caption{Quantitative comparisons on the DyDToF. }\label{tab:DyDToF}
\end{table}

\noindent \textbf{Quantitative Comparison.} 
Tabs.~\ref{tab:tanair}-\ref{tab:dynamicR} list quantitative comparisons across multiple datasets, demonstrating that our STDNet outperforms existing state-of-the-art approaches at $\times4$, $\times8$, and $\times16$ scaling factors. Specifically, Tab.~\ref{tab:tanair} shows that our method is superior to both the existing multi-frame DVSR and advanced single-frame approaches on TarTanAir. For example, compared to the second-best method, our STDNet reduces RMSE by$ 15.24cm$, MAE by $2.38cm$, and TEPE by $2.16cm$ on $\times16$ VDSR. Additionally, Tabs.~\ref{tab:DyDToF} and \ref{tab:dynamicR} further validate the generalization capability of our method on DyDToF and DynamicReplica. We observe that STDNet achieves outstanding performance, surpassing the suboptimal VDSR ($\times16$) by $4.31cm$ in RMSE on DyDToF and by $0.15cm$ in RMSE on DynamicReplica.

\begin{table}[t]
\centering
\small
\renewcommand\arraystretch{1.05}
\setlength{\tabcolsep}{3pt}{
\begin{tabular}{l|cccccc}
\toprule 
\multirow{2}{*}{Methods}   &\multicolumn{2}{c}{$\times4$}   &\multicolumn{2}{c}{$\times8$}  &\multicolumn{2}{c}{$\times16$} \\ 
\cmidrule(lr){2-3}\cmidrule(lr){4-5}\cmidrule(lr){6-7}
&RMSE$\downarrow $ &MAE$\downarrow $     &RMSE$\downarrow $ &MAE$\downarrow $  
 &RMSE$\downarrow $ &MAE$\downarrow $    \\ \midrule
FDSR                                         &0.42  &0.08                         &0.80  &0.31               &1.51  &0.66        \\
SUFT                                         &0.47  &0.17                         &0.79  &0.27               &0.56  &0.81        \\        
SGNet                                        &0.44  &0.08                         &0.72  &0.25               &1.50  &0.75        \\
C2PD                                         &0.42  &0.11                         &0.67  &0.24               &1.40  &0.65        \\
DORNet                                       &0.46  &0.10                         &0.60  &0.17               &1.39  &0.79        \\
DVSR                                         &\underline{0.37}  &\underline{0.07}                         &\underline{0.58}  &\underline{0.13}               &\underline{1.25}  &\textbf{0.48}        \\
\textbf{STDNet}                              &\textbf{0.32}  &\textbf{0.05}                         &\textbf{0.53}  &\textbf{0.10}               &\textbf{1.10}  &\underline{0.51}        \\
\bottomrule
\end{tabular}}
\caption{Quantitative comparisons on the DynamicReplica.}\label{tab:dynamicR}
\end{table}

\noindent \textbf{Visual Comparison.} Figs.~\ref{fig:TarTanAir_v} and ~\ref{fig:DyDToF_v} provide visual comparisons, clearly indicating that our method achieves more accurate depth recovery. For example, compared to previous approaches, the structure and shape of chair in Fig.~\ref{fig:TarTanAir_v} predicted by STDNet align more closely with the GT depth while exhibiting superior temporal stability. Additionally, Fig.~\ref{fig:DyDToF_v} shows that our method results in more precise reconstruction of dynamic objects (e.g., dog) than others.

\begin{figure}[t]
 \centering
 \includegraphics[width=0.85\columnwidth]{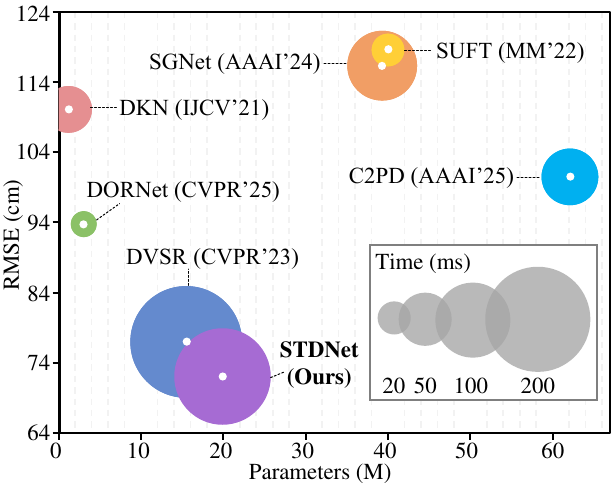}\\
 \caption{Complexity on the TarTanAir ($\times8$) tested by a 4090 GPU. A larger circle area indicates larger time.}
 \label{fig:params}
\end{figure}

\noindent \textbf{Model Complexity Analysis.} Fig.~\ref{fig:params} demonstrates that our STDNet maintains a comparable balance between parameters, performance, and inference time. Compared to single-frame approaches (DKN, SUFT, SGNet, C2PD, and DORNet), although our method exhibits higher time cost, it achieves a significant average reduction of $9.23M$ parameters and $35.82cm$ RMSE. Furthermore, STDNet outperforms the multi-frame DVSR with a $47.35ms$ improvement in inference speed and a $4.93cm$ gain in performance, while introducing only a modest parameter increase of $4.4M$.

\section{Ablation Study}
\begin{figure}[t]
 \centering
 \includegraphics[width=1\columnwidth]{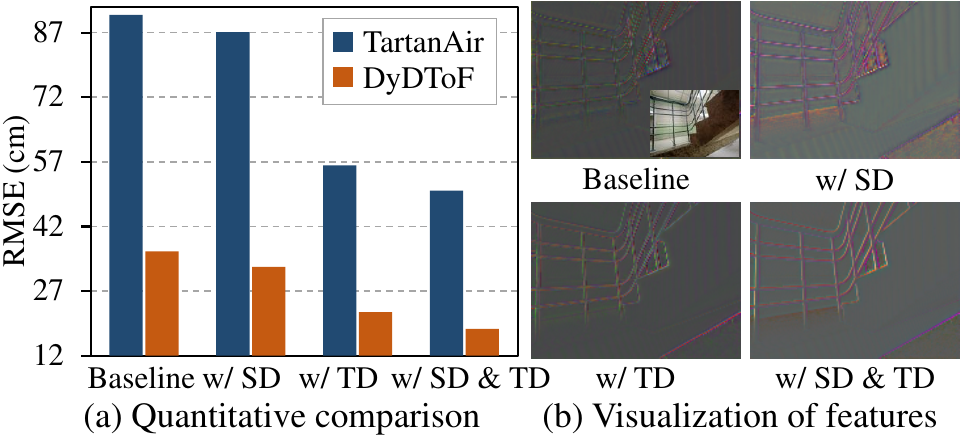}\\
 \caption{Ablation study of spatial difference (SD) and temporal difference (TD) on the TarTanAir and DyDToF ($\times4$).}
 \label{fig:ab_sd_td}
\end{figure}

\noindent \textbf{Spatial Difference and Temporal Difference.} Fig.~\ref{fig:ab_sd_td} illustrates an ablation study of spatial difference (SD) and temporal difference (TD). For the baseline, we replace all SD and TD in STDNet with concatenation operations, while keeping other architecture unchanged. As shown in Fig.~\ref{fig:ab_sd_td}(a), both SD and TD contribute to performance improvements over the baseline. For example, SD and TD reduce RMSE by $3.56cm$ and $14.02cm$ respectively on the DyDToF. When deployed together, STDNet achieves the best performance, surpassing the baseline by $17.94cm$ on the DyDToF.

Furthermore, Fig.~\ref{fig:ab_sd_td}(b) visualizes intermediate depth features using principal
component analysis (PCA). It is clearly evident that both SD and TD facilitate more accurate depth structure. When SD and TD are combined, our STDNet produces sharper and clearer depth predictions. In summary, both quantitative and visual results demonstrate that our method effectively enhances VDSR performance, reconstructing high-quality depth videos.

\begin{figure}[t]
 \centering
 \includegraphics[width=1\columnwidth]{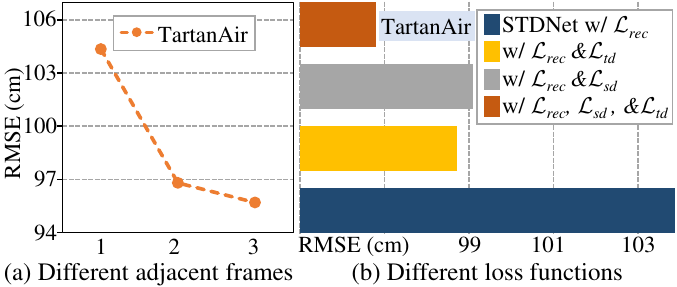}\\
 \caption{Ablation study of STDNet with (a) different numbers of adjacent frames and different loss functions ($\times16$).}
 \label{fig:ab_loss}
\end{figure}

\noindent \textbf{Different Numbers of Adjacent Frames.} Fig.~\ref{fig:ab_loss}(a) reveals the impact of different numbers of neighboring frames in TD. Compared to using only one RGB-D frames, incorporating additional neighboring frames significantly reduces the RMSE. However, the performance gains diminish when the number exceeds $2$. Consequently, to balance computational cost and performance, our STDNet selects $2$ frames (a adjacent frame and a cross frame) as the default setting.

\noindent \textbf{Different Loss Functions.} Fig.~\ref{fig:ab_loss}(b) presents the ablation study of different loss functions.  The baseline is STDNet with the reconstruction loss $\mathcal{L} _{rec}$. These quantitative results demonstrate that both the spatial difference loss and temporal difference loss significantly improve performance. When $\mathcal{L} _{sd}$ and $\mathcal{L} _{td}$ are jointly employed, our method learns accurate spatiotemporal difference representations, thereby achieving high-quality depth restoration. Finally, STDNet surpasses the baseline by $7.08cm$ in RMSE on TarTanAir.

\section{Conclusion}

In this paper, we propose the spatiotemporal difference network, a novel framework that models spatiotemporal difference representations to address the inherent long-tailed distribution problems in VDSR. Specifically, we develop a spatial difference branch that incorporates a spatial difference mechanism to selectively transfer intra-frame RGB information to depth, effectively mitigating the long-tailed effects in spatial non-smooth regions. To enhance the temporal stability of predicted depth videos, our temporal difference branch implements the proposed temporal difference strategy, prioritizing the aggregation of multi-frame RGB-D features in temporal variation areas. Furthermore, a difference regularization is introduced to facilitate accurate difference representation learning. Extensive experiments demonstrate the effectiveness and superiority of our STDNet.

\section{Acknowledgments}
This work was supported by the NSFC under Grant Nos. U24A20330 and 62361166670.

\bibliography{aaai2026}

\end{document}